\documentclass[12pt,singlespacing]{article}
\usepackage{amssymb}
\usepackage{amsfonts}
\usepackage{graphicx}
\newcommand{\R}{\mathbb{R}}

\begin{document}
\title{A Note on Local Ultrametricity in Text}
\author{Fionn Murtagh \\
Department of Computer Science \\
Royal Holloway University of London \\
Egham, Surrey TW20 0EX, England \\
E-mail fmurtagh@acm.org}

\maketitle

\begin{abstract}
High dimensional, sparsely populated data spaces have been characterized in 
terms of ultrametric topology.  This implies that there are natural, not
necessarily unique, tree or hierarchy structures defined by the 
ultrametric topology.  In this note we study the extent of local 
ultrametric topology in texts, with the aim of 
finding unique ``fingerprints'' for a text or corpus, discriminating between
texts from different domains, and opening up the possibility of 
exploiting hierarchical structures in the data.  
We use coherent and meaningful collections of 
over 1000 texts, comprising over 1.3 million words.   
\end{abstract}
%network \sep complete graph \sep edge weighted \sep metric \sep 
%Euclidean \sep ultrametric \sep chi squared metric 
%\PACS{
%{89.75.Hc}{Networks and genealogical trees} \and
%{02.50.Sk}{Multivariate analysis} \and
%{89.75.Kd}{Patterns} \and
%{89.75.Fb}{Structures and organization in complex systems}
%} % end of PACS
%}  % end of abstract

\section{Introduction}

Structures that are inherent to data of any type can be of importance, and 
hierarchical structure is a prime example.   In this work we take text
corpora and assess the extent of hierarchical structure among words 
constituting the texts.  By comprehensively taking context into account we 
seek to study hierarchical structures in the domain semantics.

The data studied in Rammal et al.\ (1986) and Murtagh (2004)  is point pattern 
data: observational features with their measurements on many coordinate 
dimensions.  Data may be instead presented as time-varying signals and 
in a similar way, related to the findings of Rammal et al.\ (1986) and 
Murtagh (2004),  
we have investigated ultrametric-related 
properties of time series or 1D signals in 
Murtagh (2005a).  In the latter time series work, we encoded the data in a 
particular way.  In this paper, we show how texts can also be 
characterized in a similar manner.

The triangular inequality holds for a metric space: $d(x,z) \leq 
d(x,y) + d(y,z)$ for any triplet 
of points $x,y,z$.  In addition the properties 
of symmetry and positive definiteness are respected.  The ``strong 
triangular inequality'' or ultrametric inequality is: $d(x,z) \leq 
\mbox{ max } \{ d(x,y), d(y,z) \}$ for any triplet $x,y,z$.  An ultrametric
space implies respect for a range of stringent properties.  For example, 
the triangle formed by any triplet is necessarily isosceles, with the two
large sides equal; or is equilateral.  Any agglomerative hierarchical 
procedure (cf.\ Benz\'ecri, 1978; Lerman, 1981; Murtagh, 1983, 1985) can 
impose hierarchical structure.  Our aim in this work is to assess 
inherent extent of hierarchical structure.  

We take a large
number of coherent collections of meaningful texts.  Through shared words,
we can define a similarity network between all texts in each of the 
collections we chose.  Aspects of the semantics of the given collection are
captured in this way.  We investigate how ultrametric each of these 
semantic networks is.  

%We select texts 
%each containing roughly 500 to 1000 words (but as will be seen below, 
%some texts had up to around 44,000 words).  
Our selected texts in this study are in English and 
do not contain accented characters (and this can be easily catered for).
These were: fairy tales by the Brothers Grimm; novels by the English 
writer, Jane Austen; in order to have very technical language, aircraft
accident reports from the US National Transport Safety Board; and in order
to seek linkages with biological and cognitive processes, a range of 
dream reports from the online DreamBank repository.  

We find clear distinctions between the semantic networks (or text collections)
studied, in terms of their relative (albeit small) extent of ultrametricity.  

Our objectives in such assessment of inherent, local, hierarchical 
structure include the following:

\begin{enumerate}
\item Ontologies (see e.g.\ G\'omez-Perez et al., 2004) have become of 
great interest to facilitate information resource discovery, and to 
support querying and retrieval of information, in current areas of work 
such as the semantic web.  Automatic or semi-automatic 
construction of ontologies is aided greatly by hierarchical relationships
between terms.  The characterizing of texts in terms of local 
hierarchical structure simultaneously provides justification for unambiguous 
local hierarchies.  (We return to this issue of ontology creation
in the Conclusion.)

\item Structures defined on terms that are more general than grammars
may be of use in modelling and assessing consistency of textual data 
(see Sasaki and P\"onninghaus, 2003); and perhaps in mapping some aspects of 
semantics and flow of reason and logic in text. 
%(for example, providing 
%a quantitative expression of Freud's concepts of
% condensation and displacement).  

\item Limited extent of hierarchical structure may point to the 
undesirability of a global tree or hierarchical clustering model for the 
text or set of texts.  However for the same reason, a set of 
local hierarchical clusterings, or a forest of (locally defined) trees, may be
more appropriate.  

We note that our work is quite different from Leo
Breiman's random forest methodology, where classification trees are
fitted multiply to a
data set.  Our work, as opposed to this, is directed towards the finding of
``shrubs'' or tree fragments in a data set.

\item Latent ultrametric distances were estimated by Schweinberger and Snijders
(2003) in order to represent transitive structures among pairwise 
relationships.  

\item Further motivation is provided by fingerprinting of authorship, and
document clustering (e.g.\ to facilitate retrieval).

\end{enumerate}

\section{Methodology}

We employ correspondence analysis for metric embedding,
followed by determination of the extent of  ultrametricity, in factor
space, based on the alpha coefficient of ultrametricity.  Our motivation 
for using precisely this Euclidean embedding is as follows.  Our input 
data is in the form of frequencies of occurrence.  Now, a Euclidean distance
defined on vectors with such values is not appropriate.  

The $\chi^2$ distance
is an appropriate weighted Euclidean distance for use with such data
(Benz\'ecri, 1979; Murtagh, 2005b).  
Consider texts $i$ and $i'$ crossed by words $j$.  Let $k_{ij}$ be the number of
occurrences of word $j$ in text $i$.  Then, omitting a constant, 
the $\chi^2$ distance between texts $i$ and $i'$ is given by 
$ \sum_j 1/k_j ( k_{ij}/k_i - k_{i'j}/k_{i'} )^2$.  The weighting term is 
$1/k_j$.  The weighted Euclidean distance is between the {\em profile} 
of text $i$, viz.\ $k_{ij}/k_i$ for all $j$, and the analogous 
{\em profile} of text $i'$.

\subsection{Alpha Coefficient of Ultrametricity}

The definition of ultrametricity introduced in Murtagh (2004) and justified 
relative to alternatives was, in 
summary, as follows.  For all triplets of points, we consider the three
internal angles.  We require that the smallest angle be less than or equal
to 60 degrees.  Then we require that the two remaining angles be
approximately equal.  Approximate equality is defined as less than 2 degrees,
in order to cater for imprecise coordinate measurement (e.g., due to 
floating point values) in an acceptable way.  Satisfying these angular 
constraints implies that the triplet of points defines an approximate 
isosceles (with small base) or equilateral triangle.  We define a
coefficient of ultrametricity of the point set as the proportion of all 
triangles satisfying these requirements.  The coefficient of ultrametricity 
is 1 for perfectly ultrametric data; and if 0 no triangle satisfies the 
isosceles or equilateral requirements.  This coefficient is 
referred to as alpha below in this article.  

As already noted, assessing ultrametricity through triangle properties 
is based on the prior  correspondence analysis, and this has the following
beneficial (and, in a sense, enabling) implications.  The correspondence
analysis 
factor space is  Euclidean.  A Euclidean space, as a particular Hilbert 
space, is a complete, normed vector space endowed with a scalar product.  
It is precisely the scalar product that allows us to define angles and
hence the triangle properties that we need.  

\subsection{Correspondence Analysis: 
Mapping $\chi^2$ into Euclidean Distances}

As a dimensionality reduction technique 
correspondence analysis is particularly appropriate for handling 
frequency data.  As an example of the latter, frequencies of word
occurrence in text will be studied below.  

The given contingency table (or numbers of occurrence) 
data is denoted $k_{IJ} =
\{ k_{IJ}(i,j) = k(i, j) ; i \in I, j \in J \}$.  $I$ is the set of text
indexes, and $J$ is the set of word indexes.  We have
$k(i) = \sum_{j \in J} k(i, j)$.  Analogously $k(j)$ is defined,
and $k = \sum_{i \in I, j \in J} k(i,j)$.  Next, $f_{IJ} = \{ f_{ij}
= k(i,j)/k ; i \in I, j \in J\} \subset \R_{I \times J}$,
similarly $f_I$ is defined as  $\{f_i = k(i)/k ; i \in I, j \in J\}
\subset \R_I$, and $f_J$ analogously.  What we have described here is 
taking numbers of occurrences into relative frequencies.

The conditional distribution of $f_J$ knowing $i \in I$, also termed
the $j$th profile with coordinates indexed by the elements of $I$, is:

$$ f^i_J = \{ f^i_j = f_{ij}/f_i = (k_{ij}/k)/(k_i/k) ; f_i \neq 0 ;
j \in J \}$$ and likewise for $f^j_I$.  

Note that the input data values here are always non-negative reals.  The 
output factor projections (and contributions to the principal directions 
of inertia) will be reals.  

\subsection{Input: Cloud of Points Endowed with the Chi Squared Metric}

The cloud of points consists of the couple: profile coordinate and mass.
We have $ N_J(I) = \{ ( f^i_J, f_i ) ; i  \in I \} \subset \R_J $, and
again similarly for $N_I(J)$.

The moment of inertia is as follows: 
$$M^2(N_J(I)) = M^2(N_I(J)) = \| f_{IJ} - f_I f_J \|^2_{f_I f_J} $$
\begin{equation}
= \sum_{i \in I, j \in J} (f_{ij} - f_i f_j)^2 / f_i f_j
\end{equation}
The term  $\| f_{IJ} - f_I f_J \|^2_{f_I f_J}$ is the $\chi^2$ metric
between the probability distribution $f_{IJ}$ and the product of marginal
distributions $f_I f_J$, with as center of the metric the product
$f_I f_J$.  Decomposing the moment of inertia of the cloud $N_J(I)$ -- or 
of $N_I(J)$ since both analyses are inherently related -- furnishes the 
principal axes of inertia, defined from a singular value decomposition.

\subsection{Output: Cloud of Points Endowed with the Euclidean 
Metric in Factor Space}

From the initial frequencies data matrix, a set of probability data,
$f_{ij}$, is defined by dividing each value by the grand total of all
elements in
the matrix.  In correspondence analysis,
each row (or column) point is considered to have an
associated weight.  The weight of the $i$th row point is given
by $f_i = \sum_j x_{ij}$, and the weight of the $j$th column point
is given by $f_j = \sum_i x_{ij}$. We consider the row points to have
coordinates ${f_{ij} / x_i}$, thus allowing points of the same
{\em profile} to be identical (i.e., superimposed). The following weighted
Euclidean distance, the $\chi^2$ distance, is then used between row
points:
$$ d^2(i,k) = \sum_j {1 \over x_j} \left( {f_{ij} \over x_i} -
                                     {f_{kj} \over x_k} \right)^2 $$
and an analogous distance is used between column points.

The mean row point is given by the weighted average of all row
points:
$$ \sum_i f_i {f_{ij} \over f_i} = f_j$$
for $j = 1, 2, \dots, m$.  Similarly the mean column profile has
$i$th coordinate $f_i$.

We
first consider the projections of the $n$
profiles in $\R^m$ onto an axis, ${\bf u}$.  This is given by
$$ \sum_j {f_{ij} \over x_i} {1 \over x_j} u_j$$ for all $i$ (note
the use of the scalar product here).  For details on determining the 
new axis, ${\bf u}$, see Murtagh (2005).

The  projections of points onto
axis ${\bf u}$ were with respect to the ${1 / f_i}$ weighted Euclidean
metric.  This makes interpreting projections very difficult from a
human/visual point of view, and so it is more natural to present results
in such a way that projections can be simply appreciated.  Therefore
{\em factors} are defined, such that the projections of row vectors
\index{factor}
onto factor ${\bf \phi}$ associated with axis ${\bf u}$ are given by
$$\sum_j {f_{ij} \over x_i} \phi_j$$ for all $i$.  Taking $$\phi_j =
{1 \over f_j} u_j$$ ensures this and projections onto ${\bf \phi}$
are with respect to the ordinary (unweighted) Euclidean distance.

An analogous set of relationships hold in $\R^n$ where the best
fitting axis, ${\bf v}$, is searched for.  A simple mathematical
relationship holds between ${\bf u}$ and ${\bf v}$, and between
${\bf \phi}$ and ${\bf \psi}$ (the latter being the factor associated
with axis or eigenvector ${\bf v}$):
$$ \sqrt{\lambda} \psi_i = \sum_j {f_{ij} \over f_i} \phi_j $$
$$ \sqrt{\lambda} \phi_j = \sum_i {f_{ij} \over f_j} \psi_i $$
These are termed {\em transition formulas}. 
 Axes ${\bf u}$
\index{transition formula}
and ${\bf v}$, and factors ${\bf \phi}$ and ${\bf \psi}$, are
associated with eigenvalue $\lambda$ and best fitting higher-dimensional
subspaces are associated with decreasing values of $\lambda$ (see Murtagh,
2005b, for further details).

\subsection{Conclusions on Correspondence Analysis and Introduction to the 
Numerical Experiments to Follow}

Some important points for the analyses to follow are -- firstly in relation 
to correspondence analysis: 

\begin{enumerate}

\item From numbers of occurrence data we always get (by design) 
a Euclidean embedding
using correspondence analysis.  The factors are embedded in a Euclidean 
metric.  

\item As seen in the previous subsection, the 
numbers of factors, i.e.\ number of non-zero eigenvalues, are
given by one less than the minimum of the number of observations studied
(indexed by set $I$) and the number of variables or attributes used 
(indexed by set $J$).  
The number of dimensions in factor space may be less than full rank
if there are linear dependencies present.  

\item In the experiments to follow in the next section, we  always 
have  $n < m$, where $n$ is number of texts or text segments, and $m$ is 
number of words.  This implies that inherent (full rank) 
dimensionality of the projected Euclidean 
factor space is $n - 1$.  

\item To assess stability of results,
in our studies we often take as input a word set given by the 
(for example, 1000) most highly ranked (in terms of frequency of 
occurrence)  words.  Thus we take $m = 1000, 2000,$ and the full 
attribute set (say, 
$m_{\rm tot}$) in each case, where the attributes are ordered in terms of 
decreasing marginal frequency.  In other words, we take the 1000 most
frequent words to characterize our texts; then the 2000 most frequent words; 
and finally all words.  Since $n < m$ it is not surprising that 
very similar results are found irrespective of the value of $m$, since
the inherent, projected, Euclidean, factor space dimensionality is the 
same in each case, viz., $n - 1$.  But we additionally find confirmation 
of stability of our results.  
 We will show quite convincingly that our results are 
characteristic of the texts used, in each case, and are in no way ``one off''
or arbitrary.  

%\item Purely as a baseline we will look at direct Euclidean pairwise 
%distances defined on $\{ k_{ij} | i = 1, 2, \dots , n; j = 1, 2, 
%\dots , m \}$.

\end{enumerate}

Some important points related to our numerical assessments below, in 
relation to data used, determining of ultrametricity coefficient, 
and software used, are as follows.

\begin{enumerate}

\item 
In line with one tradition of textual analysis associated with Benz\'ecri's
correspondence analysis (see Murtagh, 2005b) we take the unique full words and
rank them in order of importance.  Thus for the Brothers Grimm work,
below, we find: ``the'', 19,696 occurrences; ``and'',
14,582 occurrences; ``to'', 7380 occurrences; ``he'', 5951 occurrences; 
``was'', 4122 occurrences; and so on.  Last three, with one occurrence each:
``yolk'', ``zeal'', ``zest''.   

\item The alpha ultrametricity coefficient is based on triangles. Now, 
with $n$ graph nodes we have $O(n^3)$ possible triangles which is 
computationally prohibitive, so we instead sample.  The means and 
standard deviations below are based on 2000 random triangle vertex
realizations, repeated 20 times; hence, in each case, in total 40,000 
random selections of triangles.  

\item All text collections reported on below (section \ref{sectreal}) 
are publicly accessible (and web addresses are cited).  All texts were
obtained by us in straight (ascii) text format.
   
The preparation of the input data was carried out with programs of 
ours, written in C, and available at www.correspondances.info (accompanying 
Murtagh, 2005b).  The correspondence analysis software was written in
the public  R statistical software environment  
(www.r-project.org, again see Murtagh, 2005b) and is available at this same 
web address.  Some 
simple statistical calculations were carried out by us also 
in the R environment.  

\end{enumerate}

\section{Real Case Studies: Text Interrelationships Through Shared Words}
\label{sectreal}

We use in all over 900 short texts, given by short stories, or chapters,
or short reports.  All are in English.  Unique words are determined 
through delimitation by white space and by punctuation characters
with no distinction of upper and lower case.  In
all, over one million words are used in our studies of these texts.  
The study of word/text occurrences in a straightforward way, with no
truncation nor stemming nor other preprocessing, typifies a great deal
of the work of Benz\'ecri, and his journal {\em Les Cahiers de 
l'Analyse des Donn\'ees}, published by the French publisher Dunod over
three decades up to 1996.  This work of Benz\'ecri is 
discussed in detail in Murtagh (2005b).  

We carried out some assessments of Porter stemming (Porter, 1980)
as an alternative 
to use of whitespace- or punctuation-delimited words, without much 
difference.  

\subsection{Brothers Grimm}

As a homogeneous collection of texts we take 209 fairy tales of the Brothers 
Grimm (Ockerbloom, 2003), 
containing 7443 unique (in total 280,629) space- or 
punctuation-delimited words.  Story lengths were between 650 and 44,400 words.

To define a semantic context of increasing
resolution we took the most frequent 1000 words, followed by the most frequent
2000 words, and finally all 7443 words.  
%(We tested extensively the case of 
%just the 100 most frequent words also.  But in view of the texts versus 
%words dimensionality implications, viz.\ $ n > m$ here, and the slightly 
%more tricky interpretation, we deliberately do not report on these 
%results here.) 
We constructed a cross-tabulation of numbers of occurrences of
each word in each one of the 209 fairy tales.  This led therefore to a 
set of frequency tables of dimensions: $209 \times 1000,
209 \times 2000$ and $209 \times 7443$.    Through use of the $\chi^2$ 
distance between fairy tale texts, a correspondence analysis was carried out.  
From the three frequency tables, the contingency table crossing all pairs
of fairy tales could be examined; but it was far more convenient for us 
to proceed straight to the factor space, of dimension $209 - 1 = 208$.  The
factor space is Euclidean, so the correspondence analysis can be said to be
a mapping from the $\chi^2$ metric into a Euclidean metric space.

%\begin{table}
%\begin{center}
%\begin{tabular}{|crrrr|} \hline
%\multicolumn{5}{c}{209 Brothers Grimm fairy tales} \\ \hline
%Texts  &  Dim.  &    Original   &  Dim. & Factors  \\ \hline
%%209    &  100   &     0.0273    &  99   & 0.1002  \\
%209    &  1000  &     0.0324    &  208    & 0.1189  \\
%209    &  2000  &     0.0334    &  208    & 0.1083  \\
%209    &  7443  &     0.0324    &  208    & 0.1154  \\ \hline
%\end{tabular}
%\end{center}
%\caption{Coefficient of ultrametricity.  
%Original: frequencies of occurrence matrix defined on the 209 texts 
%crossed by: % 100, 
%1000, 2000, and all = 7443, words.  Euclidean distance 
%defined on each pair of texts.  Factors: factor projections resulting 
%from correspondence analysis, with Euclidean distance used between each 
%pair of texts.}
%\label{tabcorr}
%\end{table} 

\begin{table}
\caption{Coefficient of ultrametricity, alpha.  
Input data: frequencies of occurrence matrices defined on the 209 texts 
crossed by: %100, 
1000, 2000, and all = 7443, words.  
Alpha (ultrametricity coefficient) based
on factors: i.e., factor projections resulting 
from correspondence analysis, with Euclidean distance used between each 
pair of texts in factor space, of dimensionality 208.  
%The mean and standard deviations are each based on 20 realizations of 
%2000 triangles.
}
\label{tabcorrb}
\begin{center}
\setlength{\tabcolsep}{1mm}
\begin{tabular}{|crrrr|} \hline 
      &  \multicolumn{3}{c}{209 Brothers Grimm fairy tales}  &  \\ \hline
Texts & Orig.Dim. & FactorDim. & Alpha, mean & Alpha, sdev. \\ \hline 
%209  &  100      & 99   &  0.0939   &  0.0063 \\
209   &  1000     & 208  &  0.1236   &  0.0054 \\
209   &  2000     & 208  &  0.1123   &  0.0065 \\
209   &  7443    & 208  &  0.1147   &  0.0066 \\ \hline
\end{tabular}
\end{center}
\end{table} 

%The Euclidean distance was defined on the set of 209 fairy tales, based
%on the four different semantic contexts (i.e., based on characterization 
%by %100, 
%1000, 2000 and 7443 words).  

%Secondly the chi squared distance or weighted Euclidean distance between 
%profiles was used as an appropriate way to assess relative similarity. If
%$k_{ij}$ is the number of occurrences of word $k$ in text $i$, then the
%chi squared distance between texts $i$ and $i'$ is $d_\chi(i,i') =
%\sum_j k/k_j (k_{ij}/k_i - k_{i'j}/k_{i'}$ where for text $i$, $k_{ij}/k_i$
%for all words $j$ defines the text's profile; $k_i = \sum_j k_{ij}$; 
%similarly word $j$'s weight is $k_j = \sum_i k_{ij}$; and finally the 
%overall total of words in all texts is $k = \sum_i \sum_j k_{ij}$.  This 
%distance is well established for discrete data such as frequencies of 
%o%ccurence.  As can be seen, weights ($k_i$, $k_j$) are used to 
%c%ounter-balance overly frequent (or rare) words or unusually long (or 
%short) texts.  This chi squared metric is mapped into a Euclidean space
%by determining principal axes of orientation, which correspond to 
%axes of intertia, in correspondence analysis (Murtagh, 2005).  The factor
%projections will then define a Euclidean coordinate system.  It is this
%which we use, rather than the original chi squared metric, in our 
%experiments.   

%For the varying semantic resolution levels (viz., %100-, 
%1000-, 2000-, and 7443-dimensional) the inherent resolution level is not 

Table \ref{tabcorrb} (columns 4, 5) 
shows remarkable stability of the alpha ultrametricity
coefficient results, and such stability will be seen in all further results 
to be presented below.  The ultrametricity is not high for the Grimm 
Brothers' data: we recall that an alpha value of 0 means no triangle is 
isosceles/equilateral.  We see that there is very little ultrametric
(hence hierarchical) structure in the Brothers Grimm data (based on our 
particular definition of ultrametricity/hierarchy).

\subsection{Jane Austen}

To further study stories of a general sort, we use some works of the 
English novelist, Jane Austen.  

\begin{enumerate}
\item {\em Sense and Sensibility} (Austen, 1811), 
50 chapters = files, chapter lengths from 1028 to 5632 words.
\item {\em Pride and Prejudice} (Austen, 1813), 
61 chapters each containing between 683 and 5227 words. 
\item {\em Persuasion} (Austen, 1817), 24 chapters,
chapter lengths 1579 to 7007 words.
\item {\em Sense and Sensibility} split into 131 separate 
texts, each containing around 1000 words
(i.e., each chapter was split into files containing 5000 or fewer characters).
We did this to check on any influence by the size (total number of words) of
the text unit used (and we found no such influence).  
\end{enumerate}  

In all there were 266 texts containing a total of 9723 unique words.  We 
looked at the 1000, 2000 and all = 9723 most frequent words to 
characterize the texts by frequency of occurrence.

%\begin{table}
%\begin{center}
%\begin{tabular}{|crrrr|} \hline
%\multicolumn{5}{c}{266 J.\ Austen chapters or partial chapters} \\ \hline
%Texts  &  Dim.  &    Original   &  Dim.  &   Factors  \\ \hline
%%266    &  100   &     0.0409    &  99    &  0.1066  \\
%266    &  1000  &     0.0581    &  261   &  0.1521  \\
%266    &  2000  &     0.0601    &  262   &  0.1435  \\
%266    &  9723  &     0.0596    &  263   &  0.1420  \\ \hline
%\end{tabular}
%\end{center}
%\caption{Coefficient of ultrametricity.  
%Original: frequencies of occurrence matrix defined on the 266 texts 
%crossed by: %100, 
%1000, 2000, and all = 9273, words.  Euclidean distance 
%defined on each pair of texts.  Factors: factor projections resulting 
%from correspondence analysis, with Euclidean distance used between each 
%pair of texts.  Dimensionality of latter is necessarily less than $ 266 -1$,
%adjusted above for 0 eigenvalues = linear dependence.}
%\label{tabcorr2}
%\end{table} 

\begin{table}
\caption{Coefficient of ultrametricity, alpha.  
Input data: frequencies of occurrence matrices defined on the 266 texts 
crossed by: %100, 
1000, 2000, and all = 9723, words.  
Alpha (ultrametricity coefficient) based
on factors: i.e., factor projections resulting 
from correspondence analysis, with Euclidean distance used between each 
pair of texts in factor space.  
Dimensionality of latter is necessarily $ \leq 266 -1$,
adjusted for 0 eigenvalues = linear dependence. 
%The mean and standard deviations are each based on 40,000 realizations of 
%triangles.
}
\label{tabcorr2b}
\begin{center}
\setlength{\tabcolsep}{1mm}
\begin{tabular}{|crrrr|} \hline 
  & \multicolumn{3}{c}{266 Austen chapters or partial chapters} & \\ \hline
Texts & Orig.Dim. & FactorDim. & Alpha, mean & Alpha, sdev. \\ \hline 
%266  &  100      & 99   &  0.1001   &  0.0068 \\
266   &  1000     & 261  &  0.1455   &  0.0084 \\
266   &  2000     & 262  &  0.1489   &  0.0083 \\
266   &  9723     & 263  &  0.1404   &  0.0075 \\ \hline
\end{tabular}
\end{center}
\end{table} 

Table \ref{tabcorr2b}, again displaying very stable alpha values, indicates
that the Austen corpus is a small amount more ultrametric than the Grimms'
corpus, Table \ref{tabcorrb}.

\subsection{Air Accident Reports}

We used air accident reports to explore documents with very particular,
technical, vocabulary.  
The NTSB aviation accident database  
(Aviation Accident Database and Synopses, 2003)
contains information 
about civil aviation accidents in the United States and elsewhere.
We selected 50 reports.  Examples of two such reports used
by us: occurred Sunday, January 02, 2000 in Corning, AR,
aircraft Piper PA-46-310P, injuries -- 5 uninjured; occurred Sunday,
January 02, 2000 in Telluride, TN, aircraft: Bellanca BL-17-30A,
injuries -- 1 fatal.  In the 50 reports, there were 55,165 words.
Report lengths ranged between approximately 2300 and 28,000 words. The
number of unique words was 4261.

Sample of start of report 30: {\em On January 16, 2000, about
1630 eastern standard time (all times are eastern standard time,
based on the 24 hour clock), a Beech P-35, N9740Y, registered to a
private owner, and operated as a Title 14 CFR Part 91 personal
flight, crashed into Clinch Mountain, about 6 miles north of
Rogersville, Tennessee. Instrument meteorological conditions prevailed
in the area, and no flight plan was filed. The aircraft incurred
substantial damage, and the private-rated pilot, the sole occupant,
received fatal injuries. The flight originated from Louisville,
Kentucky, the same day about 1532.}

%\begin{table}
%\begin{center}
%\begin{tabular}{|crrrr|} \hline
%\multicolumn{5}{c}{50 aviation accident reports} \\ \hline
%Texts  &  Dim.  &    Original   &  Dim.  &   Factors  \\ \hline
%%50    &  100   &     0.0270    &  48    &   0.1063  \\
%50    &  1000  &     0.0407    &  48  &   0.1317  \\
%50    &  2000  &     0.0407    &  48   &  0.1212  \\
%50    &  4261  &     0.0413    &  48   &  0.1180   \\ \hline
%\end{tabular}
%\end{center}
%\caption{Coefficient of ultrametricity.  
%Original: frequencies of occurrence matrix defined on the 50 texts 
%crossed by: %100, 
%1000, 2000, and all = 4261, words.  Euclidean distance 
%defined on each pair of texts.  Factors: factor projections resulting 
%from correspondence analysis, with Euclidean distance used between each 
%pair of texts.  Dimensionality of latter is necessarily less than $ 50 -1$,
%adjusted above for 0 eigenvalues = linear dependence.}
%\label{tabcorr4}
%\end{table} 

\begin{table}
\caption{Coefficient of ultrametricity, alpha.  
Input data: frequencies of occurrence matrices defined on the 50 texts 
crossed by: %100, 
1000, 2000, and all = 4261, words.  
 Alpha (ultrametricity coefficient) based
on factors: i.e., factor projections resulting 
from correspondence analysis, with Euclidean distance used between each 
pair of texts in factor space.  
Dimensionality of latter is necessarily less than $ 50 -1$,
with an additional adjustment made for one 0-valued eigenvalue,
implying linear dependence. 
%The mean and standard deviations are each based on 40,000 realizations 
%triangles.
}
\label{tabcorr4b}
\begin{center}
\setlength{\tabcolsep}{1mm}
\begin{tabular}{|crrrr|} \hline 
  & \multicolumn{3}{c}{50 aviation accident reports} &  \\ \hline
Texts & Orig.Dim. & FactorDim. & Alpha, mean & Alpha, sdev. \\ \hline 
%50  &  100      & 48   &  0.1101   &  0.0081 \\
50   &  1000     & 48  &  0.1338   &  0.0077 \\
50   &  2000     & 48  &  0.1186   &  0.0058 \\
50   &  4261     & 48  &  0.1154   &  0.0050 \\ \hline
\end{tabular}
\end{center}
\end{table}

In Table \ref{tabcorr4b} we find ultrametricity values that are marginally
greater than those found for the Brothers Grimm (Table \ref{tabcorrb}).  It 
could be argued that the latter, too, uses its own technical 
vocabulary.   We would need to use more data to see if we can clearly 
distinguish between the (small) ultrametricity levels of these two 
corpora.

\subsection{DreamBank}

With dream reports (i.e., reports by individuals on their remembered 
dreams) we depart from a technical vocabulary, and instead raise the 
question as to whether dream reports can perhaps be considered as types
of fairy tale or story, or even akin to accident reports.  

From the Dreambank repository (Domhoff, 2003; DreamBank, 2004; Schneider
and Domhoff, 2004) 
we selected the following collections:
\begin{enumerate}
\item ``Alta: a detailed dreamer,'' in period 1985--1997, 422 dream reports.
\item  ``Chuck: a physical scientist,''  in period
1991--1993,  75 dream reports.
\item ``College women,'' in period 1946--1950,  681 dream reports.
\item ``Miami Home/Lab,''  in period  1963--1965,  445 dream reports.
\item ``The Natural Scientist,''  1939,  234 dream reports.
\item ``UCSC women,''  1996,  81 dream reports.
\end{enumerate}

To have adequate length reports, we requested report sizes of between
500 and 1500 words.  With this criterion, from (1) we obtained 118 reports,
from (2) and (6) we obtained no reports, from (3) we obtained 15 reports,
from (4) we obtained 73 reports, and finally from (5) we obtained 8 reports.
In all, we used 214 dream reports, comprising 13696 words.

Sample of start of report 100: {\em I'm delivering a car to a man --
something he's just bought, a Lincoln
Town Car, very nice. I park it and go down the street to find him -- he
turns out to be an old guy, he's buying the car for nostalgia -- it turns
out to be an old one, too, but very nicely restored, in excellent
condition. I think he's black, tall, friendly, maybe wearing overalls. I
show him the car and he drives off. I'm with another girl who drove
another car and we start back for it but I look into a shop first -- it's
got outdoor gear in it - we're on a sort of mall, outdoors but the shops
face on a courtyard of bricks. I've got something from the shop just
outside the doors, a quilt or something, like I'm trying it on, when
it's time to go on for sure so I leave it on the bench. We go further,
there's a group now, and we're looking at this office facade for the
Honda headquarters.}

With the above we took another set of dream reports, from one individual,
Barbara Sanders.  A more reliable (according to DreamBank, 2004) set of
reports comprised 139 reports, and a second comprised 32 reports.  In all
171 reports were used from this person.  Typical lengths were about 2500
up to 5322.  The total number of words in the Barbara Sanders set of
dream reports was 107,791.

%\begin{table}
%\begin{center}
%\begin{tabular}{|crrrr|} \hline
%\multicolumn{5}{c}{385 dream reports} \\ \hline
%Texts  &  Dim.  &    Original   &  Dim.  &   Factors  \\ \hline
%%385    &  100   &     0.0780    &  99    &   0.1379  \\
%385    &  1000  &     0.1122    &  384  &   0.2048  \\
%385    &  2000  &     0.1057    &  384   &  0.2137  \\
%385    &  11441  &     0.1288    &  384   &  0.1958   \\ \hline
%\end{tabular}
%\end{center}
%\caption{Coefficient of ultrametricity.  
%Original: frequencies of occurrence matrix defined on the 385 texts 
%crossed by: %100, 
%1000, 2000, and all = 11441, words.  Euclidean distance 
%defined on each pair of texts.  Factors: factor projections resulting 
%from correspondence analysis, with Euclidean distance used between each 
%pair of texts.  Dimensionality of latter is necessarily less than $ 266 -1$,
%adjusted above for 0 eigenvalues = linear dependence.}
%\label{tabcorr3}
%\end{table} 

\begin{table}
\caption{Coefficient of ultrametricity, alpha.  
Input data: frequencies of occurrence matrices defined on the 384 texts 
crossed by: %100, 
1000, 2000, and all = 11441, words.  
Alpha (ultrametricity coefficient) based
on factors: i.e., factor projections resulting 
from correspondence analysis, with Euclidean distance used between each 
pair of texts in factor space, of dimensionality $ 385 -1 = 384$.  
%The mean and standard deviations are each based on 40,000 
%realizations of triangles.
}
\label{tabcorr3b}
\begin{center}
\setlength{\tabcolsep}{1mm}
\begin{tabular}{|crrrr|} \hline 
 & \multicolumn{3}{c}{385 dream reports}  & \\ \hline
Texts & Orig.Dim. & FactorDim. & Alpha, mean & Alpha, sdev. \\ \hline 
%385  &  100      & 99   &  0.1413   &  0.0090 \\
385   &  1000     & 384  &  0.1998   &  0.0088 \\
385   &  2000     & 384  &  0.1876   &  0.0095 \\
385   &  11441    & 384  &  0.1933   &  0.0087 \\ \hline
\end{tabular}
\end{center}
\end{table} 

First we analyzed all dream reports, furnishing Table \ref{tabcorr3b}. 

In order to look at a more homogeneous subset of dream reports, we 
then analyzed separately 
the Barbara Sanders set of 171 reports, leading to Table \ref{tabcorr333b}.  
(Note that this analysis is on a subset of 
the previously analyzed dream reports, Table \ref{tabcorr3b}).  
The Barbara Sanders subset of 171 reports contained 7044
unique words in all.

Compared to Table \ref{tabcorr3b} based on the entire dream report 
collection, Table \ref{tabcorr333b} which is based on one person 
shows, on average, higher ultrametricity levels.  It is interesting to note
that the dream reports, collectively, are higher in ultrametricity level 
than our previous values for alpha; and that the ultrametricity level is 
raised again when the data used relates to one person.  

\subsection{James Joyce's Ulysses, and Overall Summary}

We carried out a study of James Joyce's {\em Ulysses}, comprising 
304,414 words in total.  We broke this text into 183 separate files, 
comprising approximately between 1400 and 2000 words each.  The number of 
unique words in these 183 files was found to be 28,649 words.  The 
ultrametricity alpha values for this collection of 183 Joycean texts 
were found to be less than the Barbara Sanders values, but higher than the 
global set of all dream reports.  
% CORRECTION WITH NEW PROGRAMS 9 MAY:  no of unique words was up from 28,631
% NEW MEAN FOR 7000 WAS: 0.2057
For 183 text segments, with frequencies of occurrence of 7000 (top-ranked)
words, we found a mean alpha of 0.2057, with standard deviation 0.0092.

%\begin{table}
%\begin{center}
%\begin{tabular}{|crrrr|} \hline
%\multicolumn{5}{c}{171 Barbara Sanders dream reports} \\ \hline
%Texts  &  Dim.  &    Original   &  Dim.  &   Factors  \\ \hline
%%171    &  100   &     0.0816    &  99    &   0.1405  \\
%171    &  1000  &     0.1212    &  170  &   0.2470  \\
%171    &  2000  &     0.1293    &  170   &  0.2110  \\
%171    &  11441  &     0.1324    &  170   &  0.2404   \\ \hline
%\end{tabular}
%\end{center}
%\caption{Coefficient of ultrametricity.  
%Original: frequencies of occurrence matrix defined on the 171 texts 
%crossed by: %100, 
%1000, 2000, and all = 7044, words.  Euclidean distance 
%defined on each pair of texts.  Factors: factor projections resulting 
%from correspondence analysis, with Euclidean distance used between each 
%pair of texts.  Dimensionality of latter is necessarily less than $ 171 -1$,
%with no adjustment necessary for 0 eigenvalues = linear dependence.}
%\label{tabcorr333}
%\end{table} 

\begin{table}
\caption{Coefficient of ultrametricity, alpha.  
Input data: frequencies of occurrence matrices defined on the 171 texts 
crossed by: %100, 
1000, 2000, and all = 7044, words.  
Alpha (ultrametricity coefficient) based
on factors: i.e., factor projections resulting 
from correspondence analysis, with Euclidean distance used between each 
pair of texts in factor space, of dimensionality $ 171 -1 = 170$. 
%The mean and standard deviations are each based on 40,000 
%realizations of triangles.
}
\label{tabcorr333b}
\begin{center}
\setlength{\tabcolsep}{1mm}
\begin{tabular}{|crrrr|} \hline 
 & \multicolumn{3}{c}{171 Barbara Sanders dream reports}  & \\ \hline
Texts & Orig.Dim. & FactorDim. & Alpha, mean & Alpha, sdev. \\ \hline 
%171  &  100      & 99   &  0.1592   &  0.0063 \\
171   &  1000     & 170  &  0.2250   &  0.0089 \\
171   &  2000     & 170  &  0.2256   &  0.0112 \\
171   &  7044     & 170  &  0.2603   &  0.0108 \\ \hline
\end{tabular}
\end{center}
\end{table} 

%Ulysses text: http://www.lib.ru/DVOJS/ulysses.txt

A summary of all our results is in Table \ref{tabsum}.  A few words of explanation
follow.  The lower values of ultrametricity can be explained by a more
common, shared word set; viz., shared over the text segment set.  The 
higher values of ultrametricity are associated with dreams, in particular 
with a single dreamer, and with {\em Ulysses}: one could argue that 
characteristics of these data sets include frequent changes in interest,
and frequent replacement of one scene, and one set of personages,
 with another.  In factor space, this implies that a triplet of points 
is  more likely to be isosceles with small base, or equilateral, compared to 
the alternative (low ultrametricity case) of more smooth transitions from 
one sentence, paragraph or section to another.   

\begin{table}
\begin{center}
\begin{tabular}{|lrrr|}\hline
Data                 &   No. texts &  No. words &  ultrametricity  \\ \hline
Grimm tales          &   209       &    7443    &     0.1147       \\
aviation accidents   &    50       &    4261    &     0.1154       \\
Jane Austen novels   &   266       &    9723    &     0.1404       \\
dream reports        &   385       &   11441    &     0.1933       \\
Joyce's Ulysses      &   183       &   28631    &     0.2057       \\
single person dreams &   171       &    7044    &     0.2603       \\ \hline
\end{tabular}
\end{center}
\caption{Summary of results for the full word set, with the exception of
the Joyce data, where 7000 words were used.  The ultrametricity is the 
alpha measure used throughout this article, where 1 is respect for 
ultrametricity by all triangles, and and 0 is non-respect in all cases.}
\label{tabsum}
\end{table}

\section{Conclusion}

We studied a range of text corpora, comprising over 1000 texts, or text
segments,  
containing over 1.3 million words.  We found very stable ultrametricity 
quantifications of the text collections, across numbers of most frequent 
words used to characterize the texts, and sampling of triplets of texts.  
We also found that in all cases (save, perhaps, the Brothers Grimm versus 
air accident reports) there was a clear distinction between the ultrametricity
values of the text collections.  

%We end with a few remarks which much remain as speculation until far more 
%sizable tests have been carried out (involving a far greater number of texts).
%However even speculation serves to motivate future work. 
Some very intriguing ultrametricity characterizations were found in our
work.  For example, we found that the technical vocabulary of air accidents 
did not differ greatly in terms of inherent ultrametricity compared to the 
Brothers Grimm fairy tales.  Secondly we found that novelist Austen's 
works were distinguishable from the Grimm fairy tales.  Thirdly we found 
dream reports to be have higher ultrametricity level than the other 
text collections.  Further exploration of these issues will require 
availability of very high quality textual data.  

Values of our alpha ultrametricity coefficient were small but 
revealing and useful nonetheless.  Ultrametricity implies hierarchical
embedding, or structuring in terms of embedded sets.  This is what we are
finding locally (and not globally) in our data. The use of such 
hierarchical fragments as relations of dominance between concepts could be
of use for ontologies.  

Ontologies, or concept hierarchies, are used 
to help the user in information retrieval in a range of ways including: 
tree-based homing in on content to be retrieved; characterizing the 
content of data repositories before querying starts; 
and disambiguating different
but overlapping content domains.  In \cite{autoonto} we explore the use
of local ultrametric embedding for ontology fragments.  As an example, 
we use Aristotle's {\em Categories} and some other modern texts (on 
ubiquitous computing, and from Wikipedia), and we 
also discuss an online web-based demonstrator supporting retrieval through 
a visual user interface.


\begin{thebibliography}{99}

\bibitem{refa1}
Austen, J. (1811).  {\em Sense and Sensibility}.  Available at: \\
http://www.pemberley.com/etext/SandS

\bibitem{refa2}
Austen, J. (1813).  {\em Pride and Prejudice}.  Available at: \\
http://www.pemberley.com/etext/PandP

\bibitem{refa3}
Austen, J. (1817).  {\em Persuasion}.  Available at: \\
http://www.pemberley.com/etext/Persuasion

%\bibitem{ref1}
%A.-L. Barab\'asi, ``Self-organized networks: resources'', at 
%www.nd.edu/$\sim$networks/database (2004).

\bibitem{ref2}
Benz\'ecri, J.P. (1979a).  {\em L'Analyse des Donn\'ees Tome 1, 
La Taxinomie}, 2nd ed., Dunod, Paris.

\bibitem{ref3}
Benz\'ecri, J.P. (1979b).  {\em L'Analyse des Donn\'ees Tome 2, 
Correspondances}, 2nd ed., Dunod, Paris.

%\bibitem{ref4}
%G. Caldarelli, A. Erzan and A. Vespignani, Eds., Special issue on Networks,
%European Physical Journal B {\bf 38}, no. 2 (2004). 

%\bibitem{ref5}
%Comtet, L. (1974).  {\em Advanced Combinatorics}, Reidel, Dordrecht.

\bibitem{ref6}
Domhoff, G.W. (2003).
{\em The Scientific Study of Dreams: Neural Networks,
Cognitive Development and Content Analysis}, American Psychological
Association.

%\bibitem{ref7}
%Donaghey, R. (1975).
%Alternating Permutations and Binary Increasing Trees,
%{\em Journal of Combinatorial Theory (A)}, 18:  141--148.

\bibitem{ref8} 
DreamBank (2004), Repository of Dream Reports, www.dreambank.net

\bibitem{gom}
G\'omez-P\'erez, A., Fern\'andez-L\'opez, M. and Corcho, O. (2004).
{\em Ontological Engineering (with Examples from the Areas of Knowledge
Management, e-Commerce and the Semantic Web)}, Springer, Berlin.

%\bibitem{ref9} 
%J.C. Gower, ``Some distance properties of latent root and vector
%methods used in multivariate analysis'',  Biometrika {\bf 53}, 325
%(1966).  % 325--328

\bibitem{ref10}
Lerman, I.C. (1981). 
{\em Classification et Analyse Ordinale des Donn\'ees},
Dunod, Paris.

\bibitem{ref11} 
Murtagh,  F. (1983).  A Survey of Recent Advances in Hierarchical 
Clustering Algorithms, {\em The Computer Journal}, 26: 
354--359.

%\bibitem{ref12} 
%Murtagh, F. (1984). 
%Counting Dendrograms: A Survey,
%{\em Discrete Applied Mathematics}, 7: 191--199. 

\bibitem{ref13}
Murtagh, F. (1985). 
{\em Multidimensional Clustering Algorithms},
Physica-Verlag, W\"urzburg.

\bibitem{ref14}
Murtagh,  F. (2004).  On Ultrametricity, Data Coding, and Computation,
{\em Journal of Classification}, 21: 167--184.

\bibitem{ref15} 
Murtagh, F. (2005a).  Identifying the Ultrametricity of Time Series, 
{\em European Physical Journal B}, 43: 573--579.

\bibitem{ref16} 
Murtagh, F. (2005b).  {\em 
Correspondence Analysis and Data Coding with Java and R},
Chapman and Hall/CRC Press, New York.  

\bibitem{autoonto}
Murtagh, F., Mothe, J. and Englmeier, K. (2007).  Ontology from local
hierarchical structure in text.  http://arxiv.org/abs/cs.IR/0701180

\bibitem{ref17} NTSB 
Aviation Accident Database and Synopses (2003), 
National Transport Safety Board,
accessible from http://www.landings.com
%/evird.acgi\$pass*59062640!\_h-www.landings.com/\_landings/
%pages/search/rep-ntsb.html 


\bibitem{ref18}
Ockerbloom, J.M. (2003). {\em Grimms' Fairy Tales}, 
http://www-2.cs.cmu.edu/$\sim$spok/grimmtmp

\bibitem{por}
Porter, M.F. (1980). An Algorithm for Suffix Stripping, 
{\em Program}, 14: 130--137.

\bibitem{ref19}
Rammal, R.,  Toulouse, G. and Virasoro, M.A. (1986). 
Ultrametricity for
Physicists, {\em Reviews of Modern Physics}, 58: 765--788.

\bibitem{sas}
Sasaki, F. and P\"onninghaus, J. (2003). 
Testing Structural Properties in Textual Data: Beyond Document Grammars, 
{\em Literary and Linguistic Computing}, 18: 89-100.

\bibitem{ref20} 
Schneider, A. and Domhoff, G.W. (2004). The Quantitative Study of Dreams, 
http://dreamresearch.net 

\bibitem{refxy}
Schweinberger, M. and Snijders, T.A.B. (2003).  Setting in Social Networks:
A Measurement Model, {\em Sociological Methodology}, 33: 307--342.

%\bibitem{ref21} 
%W.S. Torgerson, 
%Theory and Methods of Scaling (Wiley, New York, 1958).

%\bibitem{ref22} 
%C.J. van Rijsbergen, Information Retrieval, 2nd ed. 
%(Butterworths, 1979).

%\bibitem{ref23} 
%A. Trusina, S. Maslov, P. Minnhagen and K. Sneppen, 
%``Hierarchy measures in complex networks'',   
%Physical Review Letters {\bf 92}, 178702(4) (2004).

\end{thebibliography}
\end{document}